\documentclass[conference]{IEEEtran}

  	\usepackage[pdftex]{graphicx}
  	\graphicspath{{../pdf/}{../jpeg/}}
	\DeclareGraphicsExtensions{.pdf,.jpeg,.png}

	\usepackage[cmex10]{amsmath}
	\usepackage{mathabx}
	\usepackage{algorithmic}
	\usepackage{array}
	\usepackage{mdwmath}
	\usepackage{mdwtab}
	\usepackage{eqparbox}
	\usepackage{url}
        \usepackage{graphicx,subcaption}

\begin{document}

\title{\LARGE Smart Driver Monitoring Robotic System to
Enhance Road Safety : A Comprehensive Review}


 \author{\authorblockN{Farhin Farhad Riya\authorrefmark{1} , Shahinul Hoque\authorrefmark{2}, 
Xiaopeng Zhao \authorrefmark{3} and Jinyuan Stella Sun\authorrefmark{2}}
 
 \authorblockA{Department of Electrical Engineering and Computer Science, University of Tennessee\authorrefmark{1}\authorrefmark{2}}
 \authorblockA{Department of Mechanical, Aerospace and Biomedical Engineering\authorrefmark{3}}
 \authorblockA{friya@vols.utk.edu, shoque@vols.utk.edu, xzhao9@utk.edu and jysun@utk.edu}
 }

\maketitle

\begin{abstract}
The future of transportation is being shaped by technology, and one revolutionary step in improving road safety is the incorporation of robotic systems into driver monitoring infrastructure. This literature review explores the current landscape of driver monitoring systems, ranging from traditional physiological parameter monitoring to advanced technologies such as facial recognition to steering analysis. Exploring the challenges faced by existing systems, the review then investigates the integration of robots as intelligent entities within this framework. These robotic systems, equipped with artificial intelligence and sophisticated sensors, not only monitor but actively engage with the driver, addressing cognitive and emotional states in real-time. The synthesis of existing research reveals a dynamic interplay between human and machine, offering promising avenues for innovation in adaptive, personalized, and ethically responsible human-robot interactions for driver monitoring. This review establishes a groundwork for comprehending the intricacies and potential avenues within this dynamic field. It encourages further investigation and advancement at the intersection of human-robot interaction and automotive safety, introducing a novel direction. This involves various sections detailing technological enhancements that can be integrated to propose an innovative and improved driver monitoring system.
\end{abstract}

\begin{keywords} 
Road safety, Smart Driver Monitoring, Human-Robot Interaction, artificial intelligence
\end{keywords}

\IEEEpeerreviewmaketitle


\section{Introduction}
Human-Robot Interaction (HRI) has emerged as a pivotal field in the pursuit of enhancing various aspects of our daily lives. In the context of driver monitoring systems, the integration of robots introduces a dynamic dimension to the complex relationship between humans and machines. The imperative to enhance driver safety, coupled with advancements in sensor technologies and artificial intelligence, has encouraged researchers to explore novel avenues where robots seamlessly interact with drivers to monitor their behavior, cognition, and emotions. As the automobile industry progresses toward autonomous driving, understanding and optimizing the interaction between humans and robots in the driving environment becomes paramount. \cite{goodrich2007human}.

Driver monitoring systems, traditionally focused on assessing physiological parameters and driving performance, have evolved to incorporate elements of HRI to enhance their efficacy. This literature review aims to provide a comprehensive examination of the existing landscape of HRI in driver monitoring, evaluating the strengths and limitations of current systems.\cite{khan_comprehensive_survey} 
Moreover, this review aims to set the stage for future research by examining the cognitive and affecting characteristics of human drivers, the acceptability and usefulness of HRI systems, and the technological difficulties associated with them. 

As we navigate the challenges of existing research, this paper introduces innovative perspectives and proposes solutions to address identified gaps in the literature. From integrating advanced sensor technologies and machine learning algorithms to developing adaptive and personalized HRI systems, exploring human-centered design principles, and engaging in a forward-looking discourse on the future of research, our aim is to actively contribute to the ongoing conversation surrounding HRI for driver monitoring. This literature review extends an invitation to readers to join us in unraveling the complex dynamics of human-robot interaction within the driving context, laying the groundwork for safer, more intuitive, and harmonious interactions between humans and machines on the road.\cite{zhang_integrating_Year}

The contribution of this paper can be summarized as follows
\begin{itemize}
    \item This paper contributes by thoroughly exploring a range of existing studies in the field of driver monitoring systems, providing a comprehensive overview of the current state of research.
    \item A significant contribution lies in the identification and discussion of research gaps within the current landscape of driver monitoring systems. This analysis helps delineate areas where further research is needed.
    \item This paper identifies and addresses both technical and other challenges associated with driver monitoring systems, offering insights into potential hurdles and considerations in the implementation of such systems.
    \item The paper offers valuable guidance for future researchers by suggesting innovative ideas and pointing towards a direction for potential avenues of exploration and development in the field of driver monitoring systems.
\end{itemize}



\section{Methodology}
The methodological approach for this literature review involves a structured exploration of the evolution of driver monitoring systems, identification and analysis of challenges faced by existing systems, and the examination of potential improvements and innovations through the integration of robots. The review will concentrate on recent research, encompassing studies mostly conducted within the past five years. This focus aims to capture the current state and advancements in HRI for driver monitoring. The inclusion of studies will provide insights into challenges and opportunities, offering a comprehensive overview of the evolution of driver monitoring systems up to the present. To ensure academic rigor, preference will be given to peer-reviewed journals and conferences, and a variety of methodologies, including empirical studies, experimental research, and theoretical frameworks, will be considered. Recognizing the multidisciplinary nature of HRI in driver monitoring, sources from related fields such as artificial intelligence, robotics, psychology, and automotive technology will be incorporated.


\section{Existing Driver Monitoring Systems}
The domain of driver monitoring systems has experienced substantial evolution, propelled by the urgent demand for enhanced road safety and the emergence of state-of-the-art technologies. Traditional systems primarily focused on monitoring physiological parameters, such as heart rate and eye movement, have paved the way for more sophisticated solutions that integrate a holistic understanding of the driver's state. Contemporary driver monitoring systems incorporate various technologies, such as infrared cameras, steering angle sensors, and facial recognition software, enabling the detection of signs like drowsiness, distraction, or impairment, with real-time feedback and alerts for the driver. Noteworthy examples include systems tracking eye movements for assessing alertness levels and analyzing steering patterns to identify signs of fatigue \cite{mavely2017eye}. 

Various physiological sensor signals, including ECG, EOG, EEG, and Pulse Oximeter, allow clinicians to accurately assess psychological and behavioral states. \cite{begum2023intelligent} offers an overview and analysis of driver monitoring/alerting systems that leverage physiological sensor signals, serving as a valuable resource for researchers, scholars, and developers in this field by summarizing published research works and system implementations. Biological sensor signals, encompassing EEG, ECG, EMG, finger temperature, skin conductance, etc., are acquired via skin-attached electrodes, with physiological sensor signal processing playing a crucial role in their utilization within computerized systems. 

 Y. Dong \textit{et al.} in \cite{dong2011driver} reviews state-of-the-art technologies for monitoring driver inattention, classifying them into distraction and fatigue categories. It summarizes various approaches, including subjective reports, biological measures, physical measures, driving performance measures, and hybrid measures, emphasizing the effectiveness of hybrid measures in providing reliable solutions for real driving conditions while discussing nonlinear modeling techniques commonly employed in the literature.
 
\cite{verma2018deep} introduces an innovative real-time driver emotion monitoring system deployed "in the wild," utilizing face detection and facial expression analysis. Placing a camera inside the vehicle enables continuous observation of the driver's face, allowing regular monitoring of their emotional state. This non-intrusive approach, assessing the driver's attentiveness in naturalistic driving conditions, is crucial for an automated driver assistance system (ADAS). The system employs a robust face detection model based on a mixture of trees with a shared pool of parts, ensuring accuracy across diverse environmental conditions. Additionally, facial landmark points are extracted to enhance the emotion recognition system, incorporating convolutional neural networks in the proposed approach.

T. Brandt \textit{et al.} in \cite{brandt2004affordable} introduces a visual driver surveillance system that monitors both the driver's head motion and eye blink patterns to detect signs of fatigue and monotony. The system stands out for its use of standard equipment, achieving a cost-effective performance with fast computation, suitability for measurements in darkness, and consideration of monotony, demonstrated successfully under various conditions.

 Another work \cite{koesdwiady2017recent} discusses the increasing complexity of driving in busy environments and the essential factors for safe driving, emphasizing the importance of attentive drivers, perception, awareness, and decision-making. It provides an overview of driver safety monitoring systems, exploring sources of driver inattention, safety systems addressing this issue, and the integration of next-generation driver monitoring systems in the context of the Internet of Cars, promoting an environment where smart cars collect and utilize information for enhanced safety. The conclusion highlights ongoing issues and emerging trends in the research community.

 A. Alamri \textit{et al.} introduces a bio-signal-driven system designed for real-time identification of aggressive driving behaviors. \cite{alamri2023effective} This system employs a deep convolutional neural network (DCNN) model, utilizing both edge and cloud technologies. It comprises three modules: an in-vehicle driving behaviors detection module on edge devices, a cloud-based training module, and an analyzing module situated in a monitoring environment linked through a telecommunication network. 
 
 In \cite{yoo2023optimization} the authors propose an optimization algorithm to simplify Driver Monitoring Systems (DMS) for real-time implementation by utilizing camera images to monitor driver activities. The proposed system extracts driver state information, including head pose and gaze, from the region of interest in the camera images, demonstrating accurate detection of the driver's state and immediate warning in case of any mistakes.

In terms of detecting and analyzing face structures and objects within a frame, \cite{kunani2021realtime} project utilizes the Viola Jones and YOLO algorithms for face and object detection. The system incorporates PERCLOS to calculate the Eye Aspect Ratio (EAR) and promptly alerts the driver to instances of drowsiness or the presence of objects in real-time, showcasing a high level of accuracy in both detection aspects.
 
 Particularly noteworthy is the prevalence of deep learning models in contemporary decision-making processes. These models leverage facial recognition technologies to discern intricate details about the driver's state, facilitating the identification of signs such as tiredness and fatigue. This cutting-edge technology contributes to heightened safety measures and timely interventions. The integration of sensor data, machine learning, and deep learning techniques signifies a transformative shift in driver monitoring, providing a more nuanced understanding of human behavior in the driving context.

 



\section{Relevant Domains for Driver Monitoring System}

In this section, we thoroughly explore various domains that play a crucial role in advancing driver monitoring systems. By examining existing applications and advancements in fields such as artificial intelligence, computer vision, bio-metrics, and sensor technology, we identify opportunities for collaboration, paving the way for a more sophisticated and effective driver monitoring system. These domains collectively form a robust foundation for an innovative driver monitoring system, poised to bridge existing research gaps. Moreover, by combining the progress from different areas, the proposed new system has the power to make roads much safer. It focuses on important aspects of how drivers behave and pay attention. This section also can guide us in grasping the process of blending technologies to create a robotic system capable of effectively interacting with humans, advancing the future of road safety technology.

\subsection{Innovative Sensor Technologies}

In driver monitoring systems multi-sensor approaches including vision, audio, and physiological sensors, provides a comprehensive data-set for assessing the driver's mental and emotional states.

In Facial Expression Analysis, vision sensors, commonly embodied as high-resolution cameras play a pivotal role. These cameras, which work in harmony with powerful computer vision algorithms, capture and methodically analyze diverse face expressions. This procedure gives the system the ability to recognize a wide range of emotions, including but not limited to happy, sadness, rage, and exhaustion. The combination of high-quality vision sensors and cutting-edge algorithms improves the system's ability to perceive and interpret the subtle and dynamic language conveyed by facial expressions. This holistic approach not only makes emotion recognition easier, but it also adds to a better understanding of human interaction and behavior in the context of technology \cite{kolakowska2020review} .

Audio sensors, often in the form of microphones, are employed for Voice Tone Recognition. These sensors capture the driver's vocal nuances, including changes in tone, pitch, and speech patterns. Advanced processing algorithms then analyze this audio data to detect signs of stress, fatigue, or emotional distress, contributing valuable insights to the overall assessment by using technology known as automatic speech recognition (ASR) \cite{kolakowska2020review} .

Physiological sensors, encompassing wearable or non-intrusive devices like electrocardiography (ECG), electroencephalography (EEG), and galvanic skin response sensors, play a crucial role in gaining a comprehensive understanding of the driver's physiological responses. ECG sensors closely monitor heart rate variability, offering insights into the cardiovascular dynamics associated with stress and cognitive load. EEG sensors analyze brainwave patterns, providing valuable information about the driver's cognitive engagement and mental states during driving. Galvanic skin response sensors measure changes in skin conductance, a key indicator of emotional arousal and stress levels. By collectively examining these physiological indicators, driver monitoring systems can derive a nuanced understanding of the driver's cognitive load, stress responses, and emotional states. This multifaceted physiological assessment contributes to a more holistic evaluation of the driver's well-being and alertness, enhancing the system's overall effectiveness in promoting safe and adaptive driving experiences \cite{begum2023intelligent}.

Eye-tracking technology uses specialized vision sensors that are deliberately focused on the driver's eyes, either using infrared sensors or cameras for accurate monitoring. In real-time, these sensors capture minute information of the driver's eye movements and blink patterns. The system can undertake a full assessment of the driver's physiological and cognitive states by methodically tracking the direction of gaze, the duration and frequency of blinks, and other ocular activities. The device may assess levels of drowsiness, weariness, and overall concentration by analyzing eye movements. Long durations of fixed gazing or erratic eye movements may signal the onset of drowsiness, whereas quick or inconsistent blinking patterns may indicate exhaustion. Furthermore, the device can analyze the driver's concentration on the road, defining attentiveness levels \cite{mavely2017eye}.

Steering and lane-keeping sensors, intricately integrated into both the steering wheel and the lane-keeping assist systems, play a pivotal role in the driver monitoring system. These sensors continuously collect and analyze data pertaining to steering patterns and any deviations from the designated lane. By monitoring the driver's steering behavior, the system can effectively detect signs of inattention or drowsiness. In \cite{pohl2006driverdistraction} to enhance lane-keeping support, an additional torque is applied to the steering shaft. This torque serves to guide the vehicle back to an appropriate lane position in instances where it has veered off course. Importantly, this intervention occurs selectively, activating only when the vehicle has drifted out of its lane and the driver is identified as distracted. By applying torque to the steering mechanism, the system ensures a proactive response to potential instances of inattention, providing the driver with corrective assistance and contributing to overall road safety.

Embedded pressure sensors in the driver's seat play a crucial role in gathering information about alterations in posture and pressure distribution. This data provides additional insights into the driver's comfort and alertness levels. Theoretically, analyzing body pressure distribution allows for the estimation of driver distraction and fatigue levels, providing a means to detect changes in posture and behavior. Research suggests that driver posture is linked to driving comfort, and any decrease in comfort may lead to driver distraction. Moreover, there exists a correlation between the driver's wakefulness, fatigue levels, and their driving posture. Drivers tend to readjust or tilt back and forth more frequently as drowsiness and fatigue increase. A proposed study \cite{takamatsu2022urethane} involves the incorporation of urethane-foam-embedded silicon pressure sensors, equipped with a stress-concentration packaging structure, into the car seat. This integration aims to monitor the driver's cognitive state, posture, and driving behavior for a comprehensive understanding of their well-being during the driving experience.

Depth sensors are crucial for constructing three-dimensional representations of visible scenes by capturing depth information. Initially utilized in 3D scanners, autonomous robots, and gaming consoles like Microsoft Kinect, these sensors have now found applications in diverse fields, including autonomous vehicles, drones, laptops, and smartphones. In smartphones, these sensors can be placed on both the front and back, with rear sensors often possessing an extended range for functions like focusing, foreground object detection, and achieving a dependable depth of field and background blur (bokeh) effect. Rear depth sensors are also employed in augmented and mixed reality, 3D photography, and night vision. Depth sensors operate on various technologies, including passive stereo-photogrammetry with two optical cameras, infrared structured light pattern projection, and the prevalent time-of-flight (ToF) cameras. A variant of ToF is Light Detection and Ranging (LiDAR), which generally enables more accurate and faster depth measurements. It plays a crucial role in advancing driver monitoring systems by providing precise and real-time depth information. By emitting laser beams and measuring their return times after hitting objects, LiDAR enhances depth perception within the vehicle, enabling accurate tracking of the driver's position and movements. This detailed depth data facilitates effective gesture recognition, capturing subtle hand movements for control or communication purposes. Additionally, LiDAR contributes to precise facial and body tracking, offering insights into the driver's alertness, emotions, and overall state. The technology enables the creation of a three-dimensional map of the vehicle's interior, supporting the monitoring of the driver's posture, position, and interactions with vehicle controls. In terms of safety, LiDAR aids in drowsiness detection by capturing subtle head movements and eye blinks, allowing for timely alerts to prevent potential accidents. Furthermore, LiDAR contributes to collision avoidance by continuously scanning the vehicle's surroundings, detecting potential obstacles, and providing crucial data for collision warning systems. Unlike some traditional camera-based systems, LiDAR operates effectively in various lighting conditions, ensuring reliable performance in different driving environments. Overall, LiDAR enhances driver monitoring systems with its ability to provide accurate and versatile depth information \cite{kolakowska2020review}.

As the field of HRI for driver monitoring progresses, the integration of innovative sensor technologies stands as a cornerstone for advancing the capabilities of these systems. The subsequent sections will explore the potential of advanced data analytics, machine learning, and adaptive HRI systems, contributing to a comprehensive approach for addressing the challenges and unlocking new possibilities in driver monitoring.

\subsection{Advanced Data Analytics and Machine Learning}
In the dynamic aspects of HRI for driver monitoring, leveraging advanced data analytics and machine learning techniques offers a pathway to unlocking deeper insights and improving the responsiveness of these systems.

Using machine learning algorithms for recognizing behavioral patterns empowers the system to understand and predict the actions of the driver. This involves foreseeing alterations in driving behavior, like unexpected accelerations or deviations from the usual, to proactively evaluate the driver's condition. By employing longitudinal data analysis, personalized driver profiles can be established. Machine learning models have the capability to learn from past data, gaining insights into individual driving habits, preferences, and responses. This contributes to the development of a more flexible and personalized HRI system \cite{hsu_machine_learning_HRI} \cite{wang_machine_learning_HRI_ITS}. Moreover, machine learning models can be trained to dynamically evaluate the driver's cognitive load in real-time by incorporating features such as eye movement patterns, physiological signals, and vehicle telemetry data. This integration enables the system to adapt its interactions based on the driver's mental workload. The exploration of online learning algorithms further facilitates continuous model adaptation, ensuring that the HRI system can update its understanding of the driver's cognitive state as new data becomes available, thereby ensuring ongoing accuracy \cite{jung2023vectorized}. These models utilize algorithms such as Support Vector Machines, Random Forest, Convolutional Neural Networks, and Recurrent Neural Networks to classify and recognize behavioral patterns in driving. In assessing the driver's cognitive load in real-time, regression models like Linear Regression and Support Vector Regression, as well as ensemble methods like AdaBoost and Random Forest, are employed. The system adapts its interactions based on features such as eye movement patterns, physiological signals, and vehicle telemetry data. Additionally, online learning algorithms, including Online Gradient Descent and Adaptive Learning, facilitate continuous model adaptation, ensuring that the HRI system can update its understanding of the driver's cognitive state with ongoing accuracy as new data becomes available.

The combination of Deep Neural Networks (DNN) and Natural Language Processing (NLP) appears as a critical technique for recognizing and responding to the driver's emotional state in the context of incorporating emotional intelligence into HRI \cite{zhang2024deep}. The use of DNNs, specifically Convolutional Neural Networks (CNNs), becomes essential for the detailed analysis of face expressions. This application allows for a thorough examination of photos or video streams containing the driver's face, assuring correct emotion recognition. Simultaneously, NLP approaches, presumably based on recurrent neural networks (RNNs) or transformer models, are used to interpret audio data, extracting sentiment and emotional nuances from the driver's voice. \cite{williams-etal-2018-recognizing} DNNs contribute to the analysis of physiological information, extracting relevant elements that improve the system's ability to recognize emotions. Machine learning techniques are used in the holistic emotional computing system, which could include ensemble approaches or reinforcement learning. This dynamic combination allows for real-time emotional feedback adjustment in response to the driver's particular reactions. Because of the adaptive nature of this technique, each contact is individualized, with emotional reactions tailored to the driver's unique emotional requirements and preferences. This, in turn, improves the HRI system's overall sympathetic quality, allowing a more detailed and responsive human-machine relationship.

These approaches attempt to improve the capabilities of HRI systems for driver monitoring as the sector embraces the promising prospects of modern data analytics and machine learning. The subsequent sections will delve into adaptive and personalized HRI systems, emphasizing human-centered design principles, and addressing the overarching research gaps to shape the future of HRI in the driving environment.

\subsection{Adaptive and Personalized HRI}
Recognizing the unique nature of each driver and the dynamic nature of driving scenarios, the development of adaptive and personalized HRI systems is essential for maximizing effectiveness and user acceptance.
\subsubsection{Dynamic User Profiles}
 Implementing systems that continuously monitor user behavior allows for the creation of dynamic user profiles. Machine learning algorithms can analyze real-time data to adapt the HRI system based on the driver's changing preferences, habits, and responses. \cite{rossi_user_profiling_HRI}
Incorporating context awareness into user profiles enables the HRI system to customize interactions based on the specific driving scenario. Adapting to factors such as traffic conditions, weather, and time of day ensures that the system provides relevant and timely assistance.

\subsubsection{Adaptive Feedback Mechanisms}
Developing algorithms that dynamically adjust the timing and nature of feedback responses ensures that the HRI system is responsive to the driver's immediate needs. This includes the ability to delay non-urgent alerts during periods of high cognitive load. \cite{sekmen_assessment_adaptive_HRI}
Introducing modulated feedback intensity allows the system to calibrate the strength of alerts based on the driver's sensitivity. Understanding individual thresholds for distraction or stress enables a personalized approach to feedback delivery.

\subsubsection{Customizable Interaction Modalities}
Offering drivers the ability to define their preferences for interaction modalities ensures a personalized HRI experience. \cite{rossi_preface_special_issue} Whether through voice commands, haptic feedback, or visual cues, tailoring the system to individual preferences enhances user satisfaction.
Integrating multimodal interfaces allows drivers to choose the combination of interaction modes that suits their preferences. For example, a driver may prefer visual cues during low cognitive load but opt for haptic feedback during high-stress situations.

\subsubsection{Long-Term Learning and Adaptation}
Implementing reinforcement learning mechanisms enables the HRI system to adapt over time based on user feedback and the effectiveness of previous interactions. This continual learning process ensures ongoing improvement in personalization.
Building on the concept of continuous user monitoring, longitudinal profiling involves tracking changes in driver behavior over extended periods. This holistic understanding allows the system to adapt to long-term changes in the driver's preferences and cognitive capabilities.

\subsubsection{User-Centric Intervention Strategies}
Allowing users to set their own warning thresholds for specific events, such as lane departure or following distance, caters to individual risk tolerances. This empowers drivers to customize the level of intervention according to their comfort. \cite{rossi_preface_special_issue}
Enabling drivers to express their preferences for the type and frequency of assistance ensures a user-centric approach. This could include preferences for proactive advice, minimal intervention, or specific types of alerts. This kind of customization can also help building trust on the robotic systems which is essential for developing a successful system. \cite{schaefer_measuring_trust_HRI}

The incorporation of adaptive and personalized features into HRI systems not only enhances user satisfaction but also contributes to the overall effectiveness of driver monitoring. The subsequent sections will delve into the importance of human-centered design principles in shaping these interactions and address research gaps for a comprehensive understanding of the field.

\subsubsection{Human-Centered Design Principles}
In creating systems where robots interact with humans for driver monitoring, it is essential to follow design principles that focus on the needs and experiences of people. The integration of innovative solutions based on these principles not only enhances the usability of the system but also fosters user acceptance and trust.

Employing intuitive and user-friendly interfaces reduces cognitive load and facilitates seamless interaction. Visual displays, voice prompts, and haptic feedback should be designed with clarity and simplicity to ensure easy comprehension during driving \cite{kim_user_centered_HRI} \cite{rudigkeit_user_friendly_AHRS_HMI}.
Enhancing the system's natural language processing capabilities enables more human-like interactions. This includes the ability to understand and respond to conversational cues, making the HRI system a more integrated and familiar part of the driving experience \cite{kilicaslan_nlp_based_HRI}.

Implementing interfaces that provide clear explanations of the system's decisions enhances transparency. Drivers should have access to information about why certain alerts were triggered, promoting a sense of understanding and trust.
Allowing users to access and review their own driving data contributes to transparency. Providing insights into how the HRI system interprets their behavior fosters a collaborative relationship between the driver and the technology.

Facilitating shared decision-making between the driver and the HRI system involves presenting options and recommendations rather than enforcing strict directives. This collaborative approach empowers the driver to make informed choices \cite{azhar_study_shared_decision_making} \cite{unhelkar_decision_making_collaborative_tasks}.
Providing mechanisms for users to initiate interactions with the HRI system ensures a responsive and non-intrusive experience. This could involve a designated button for requesting information or assistance when needed.

Implementing gradual familiarization strategies introduces users to the HRI system's capabilities over time. Progressive exposure allows drivers to become acclimated to the technology, reducing potential resistance. Incorporating trust-building measures, such as clear communication of system reliability and accuracy, contributes to user confidence. Establishing trust is crucial for the successful adoption of HRI systems in the driving context.

\begin{figure}
 \begin{subfigure}{0.24\textwidth}
     \includegraphics[width=\textwidth]{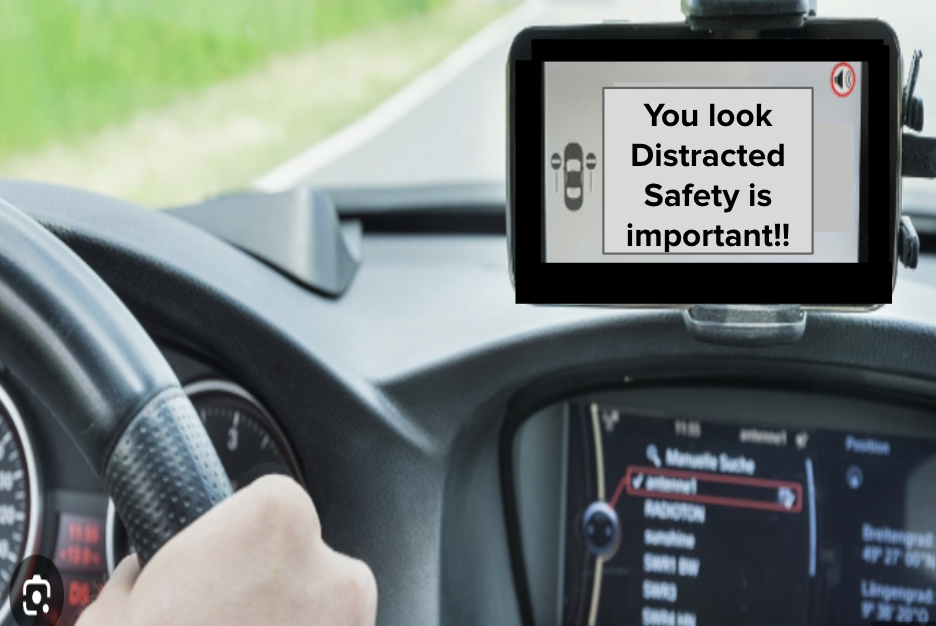}
     \caption{Robot displaying alerts for the driver \newline }
     \label{fig:a}
 \end{subfigure}
 \hfill
 \begin{subfigure}{0.24\textwidth}
     \includegraphics[width=\textwidth]{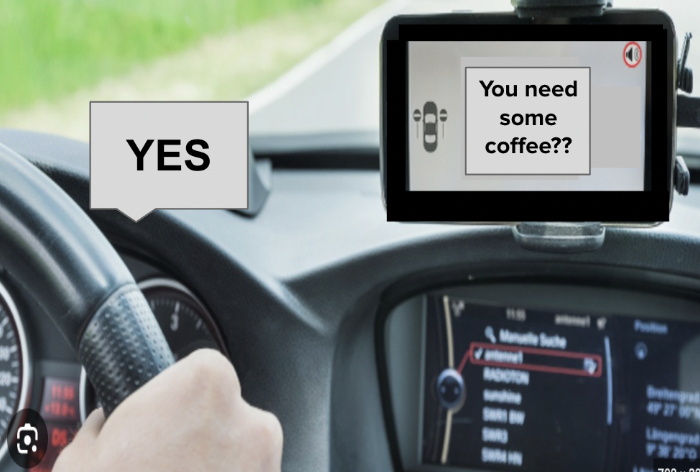}
     \caption{Robot asking the driver if he needs some coffee when detects drowsiness}
     \label{fig:b}
 \end{subfigure}
 
 \medskip
 \begin{subfigure}{0.24\textwidth}
     \includegraphics[width=\textwidth]{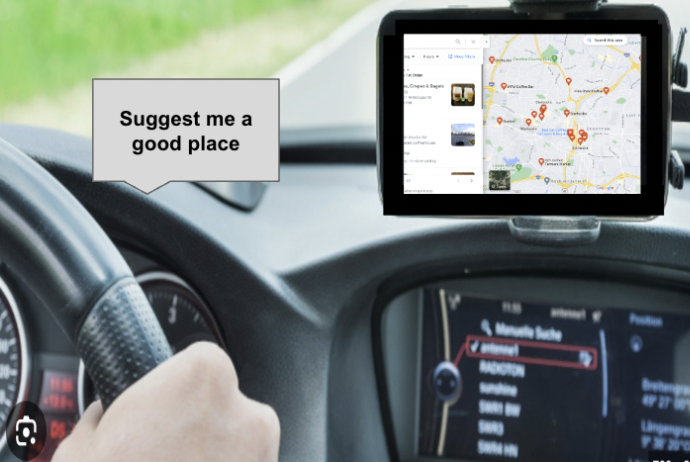}
     \caption{Robot suggesting nearby coffee shops to the driver \newline}
     \label{fig:c}
 \end{subfigure}
 \hfill
 \begin{subfigure}{0.24\textwidth}
     \includegraphics[width=\textwidth]{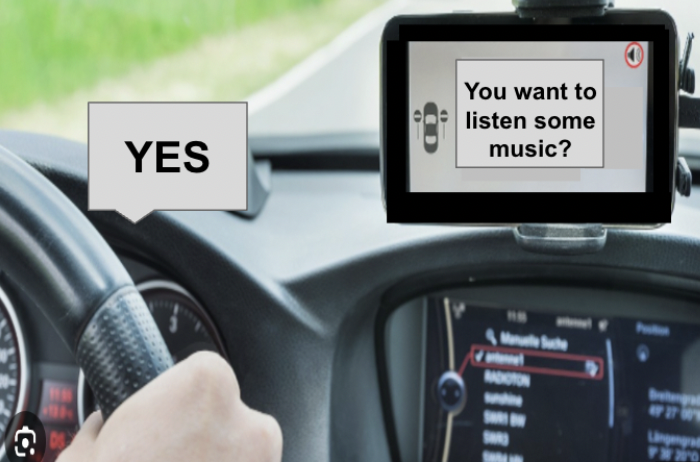}
     \caption{Robot playing music that helps the driver to concentrate in driving}
     \label{fig:d}
 \end{subfigure}

 \caption{Proactive interaction of the Robotic system with the driver}
 \label{Label}

\end{figure}

Integrating redundancy and fail-safe mechanisms ensures the system can respond effectively in case of errors or malfunctions. This safety-centric design is essential for maintaining a secure driving environment. \cite{woodman_novel_approaches_HRI_safety} Designing emergency protocols with a focus on user-centered principles involves considering the user's emotional state during crisis situations. Providing clear and empathetic guidance enhances user coping mechanisms in emergencies. \cite{wang_multimodal_HRI_smart_manufacturing}

Establishing continuous feedback loops with users allows for ongoing refinement of the HRI system \cite{wrede_analyses_of_feedback_HRI} \cite{carlmeyer_towards_closed_feedback_loops_HRI}. Regular user input ensures that updates align with evolving user preferences and needs. Adopting iterative design processes involves refining the HRI system based on real-world user experiences \cite{karami_adaptive_artificial_companions}. This iterative approach allows for the identification and correction of design flaws over successive development cycles \cite{kothig_connecting_humans_robots_HRI}.

As HRI systems for driver monitoring evolve, prioritizing human-centered design principles contributes to the creation of systems that are not only technically proficient but also genuinely aligned with user needs and expectations. The subsequent sections will explore research gaps and propose future directions for advancing HRI in driver monitoring, shaping the trajectory of this dynamic field.

\section{Proposed Robotic System in Driver Monitoring}
This section discusses the proposed idea of a robotic system in driver monitoring systems, enabling effective interactions with humans. The system exhibits the capacity to perform both proactive and active tasks, ensuring driver safety and enhancing overall road safety. The fusion of robots into driver monitoring systems represents a paradigm shift, ushering in a new era of interactive and intelligent vehicles. Robots, equipped with artificial intelligence and advanced sensor technologies, play a pivotal role in augmenting the capabilities of traditional driver monitoring systems. Their integration introduces a dynamic layer of interaction that extends beyond passive monitoring to actively engaging with the driver in real-time.

Robotic systems in driver monitoring can take various forms, from virtual assistants with natural language processing capabilities to physical robotic interfaces within the vehicle. These entities are designed to interpret and respond to the driver's cognitive and emotional states, thereby enhancing the overall effectiveness of monitoring. For instance, a robot could provide verbal prompts to a drowsy driver, or use facial recognition to assess stress levels and adjust the vehicle's environment accordingly. The literature reveals a growing body of research exploring the efficacy of integrating robots into driver monitoring, highlighting the potential for improved safety outcomes and a more intuitive driving experience. However, challenges persist, including the need for seamless human-robot collaboration, addressing user acceptance issues, and ensuring the ethical use of such technology. As we navigate through these studies, it becomes evident that the integration of robots opens up exciting possibilities, demanding further exploration and refinement to realize its full potential in enhancing driver monitoring systems.

\subsection{Proactive Interaction}

In the context of proactive interaction, the robotic system responds dynamically to instances of driver distraction, drowsiness, or fatigue, initiating a sequence of interventions. As depicted in Figure \ref{fig:a}, the robot employs visual and auditory alerts designed to capture the driver's attention and redirect focus. Transitioning to Figure \ref{fig:b}, when the system identifies signs of drowsiness, it actively engages with the driver, offering a potential solution by asking if they would like a cup of coffee. Moving forward to Figure \ref{fig:c}, upon the driver expressing interest in coffee, the robot takes a further step by suggesting nearby coffee shops, providing practical recommendations. Finally, in Figure \ref{fig:d}, recognizing the persistence of driver drowsiness, the robot proactively proposes an alternative remedy by inquiring if the driver would prefer to listen to music, offering an additional sensory stimulus to counteract fatigue. This comprehensive and responsive approach showcases the system's adaptability, offering tailored solutions to address various levels of driver distraction and fatigue in real-time scenarios.

\begin{center}
    \includegraphics[width=.95\linewidth]{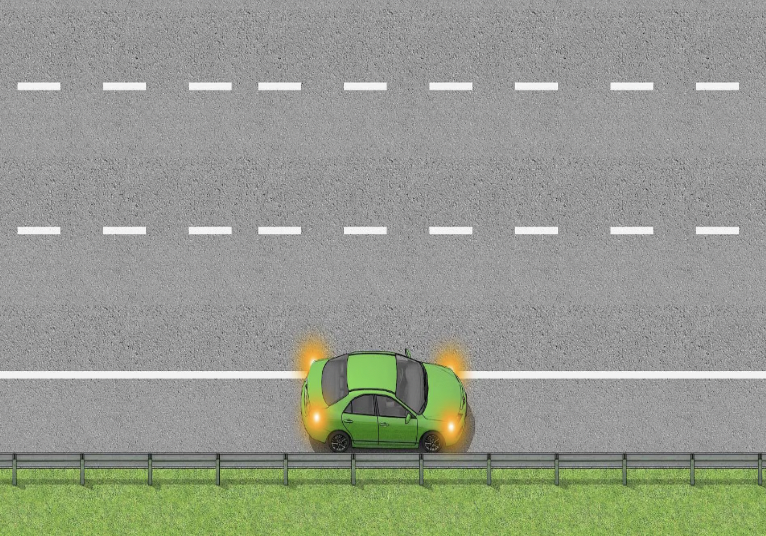}
    \captionof{figure}{Active interaction of the Robotic system with the driver}
    \label{hri5}
\end{center}

\subsection{Active Interaction}
In the active interaction scenario, the robot responds to emergency situations. When the system recognizes a crucial situation where the driver is consistently sleeping or unable to operate the vehicle, the robot takes over control of the car. In a decisive and swift action aimed at ensuring safety, the robot actively guides the vehicle to the shoulder of the road, strategically preventing potential accidents that might arise from the driver's inability to drive. Concurrently, the robotic system activates the emergency lights of the vehicle, enhancing its visibility to other road users. Additionally, the robot has the capability to initiate a call for assistance, potentially reaching out for help from roadside services or emergency responders. This active and comprehensive intervention highlights the system's capacity to assume control and implement emergency procedures swiftly when the well-being of the driver and others on the road is at risk. The visual representation of this emergency response scenario is depicted in Figure \ref{hri5}.

\section{Key Challenges in Human-Robot Interaction for Driver Monitoring}
\subsection{Cognitive and Emotional Aspects}
The successful integration of robots into driver monitoring systems hinges on their ability to discern and appropriately respond to the cognitive and emotional states of drivers. Numerous studies have explored the potential of robots to detect signs of driver fatigue, stress, and distraction. These investigations often employ a variety of sensors, including facial expression analysis, eye-tracking, and physiological sensors, to infer the driver's mental and emotional state.

\begin{itemize}
\item Driver Fatigue Detection: Several researchers have investigated the use of machine learning algorithms to analyze driving behavior and physiological signals for early detection of driver fatigue. Challenges in accurately identifying the onset of fatigue and distinguishing it from momentary lapses in attention are areas of ongoing research. \cite{sikander2023driver}

\item Cognitive Load Assessment: Understanding the cognitive load on a driver is crucial for designing effective human-robot interactions. Studies have explored methods such as workload estimation based on physiological measures and eye movement patterns to gauge the mental demand imposed by driving tasks.

\item Emotion Recognition: Recognizing and interpreting a driver's emotions is a complex task. \cite{zhangdriver} While facial expression analysis and physiological signals can provide insights, there are challenges in achieving real-time and accurate emotion recognition, particularly in dynamic driving scenarios.

\item Stress and Anxiety Detection: Identifying stress and anxiety levels in drivers is essential for proactive intervention. However, the variability in individual responses and the influence of external factors on stress make it a challenging aspect to model accurately. \cite{elgendi_machine_learning}
\end{itemize}

\subsection{Usability and Acceptance}
While the integration of robots into driver monitoring systems holds promise for enhancing safety and performance, challenges arise in ensuring the usability of such systems and gaining user acceptance within the driving context. \cite{gillan_usability_2020} presents a review on usibility issues in human-robot systems.
\begin{itemize}

\item User Interface Design: The design of human-robot interfaces within the vehicle plays a critical role in usability. Research has emphasized the need for intuitive and non-distracting interfaces that convey information effectively without compromising the driver's attention on the road.

\item Interaction Modalities: Investigating the most suitable interaction modalities is essential. Voice commands, haptic feedback, and visual displays are among the options, each posing unique challenges. Understanding the trade-offs between modalities in terms of attention demands and user preferences is an ongoing area of study.

\item Driver Attitudes and Perceptions: Assessing how drivers perceive and accept the presence of robots in the driving environment is a multifaceted challenge. Studies delve into driver attitudes toward automation, trust in robot-assisted systems, and concerns related to privacy and data security.

\item Long-Term User Engagement: Sustaining user engagement with HRI systems over time is crucial. While initial acceptance may be positive, maintaining user interest and preventing disengagement or distrust over extended periods pose challenges that necessitate longitudinal studies.

\item Dynamic Driving Environments: The unpredictable nature of real-world driving introduces challenges in designing systems that remain effective across diverse conditions. Adapting to various driving scenarios, traffic conditions, and driver behaviors requires robustness in HRI design.

\item Integrating Feedback Responsively: Providing feedback that is timely and contextually appropriate is challenging, particularly when dealing with rapidly changing driving situations. The literature reflects efforts to balance the need for immediate response with the importance of avoiding unnecessary distractions.

\item User-Centered Design Approaches: Emphasizing user-centered design principles is vital. Involving end-users in the design process and considering user preferences and feedback can enhance the acceptance of HRI systems.

\item Education and Familiarization: Addressing driver concerns and building trust may involve educational campaigns to familiarize users with the capabilities and limitations of HRI systems. Understanding how to communicate the benefits of these systems without inducing complacency is a nuanced challenge.

\end{itemize}

As researchers strive to develop effective HRI systems for driver monitoring, tackling the challenges of usability and acceptance emerges as a critical aspect. The subsequent sections will delve into the technical challenges associated with HRI implementation and propose innovative solutions to advance the field.

\subsection{Technical Challenges}
The integration of robots into driver monitoring systems introduces a set of technical challenges that must be addressed to ensure the effectiveness and reliability of HRI in the driving context.

Incorporating a diverse array of sensors, such as cameras, LiDAR, and physiological sensors, raises challenges in effectively fusing information from different modalities. Ensuring accurate and real-time integration is crucial for providing a comprehensive understanding of the driver's state. Adapting HRI systems to diverse environmental conditions, including varying lighting, weather, and road conditions, poses a significant technical challenge. Ensuring sensor accuracy and reliability under these dynamic circumstances is essential for real-world applicability. Achieving low-latency processing of sensor data is imperative for timely HRI responses. Delays in processing can impact the system's ability to provide timely feedback, particularly in situations where quick intervention is essential for safety. Exploring solutions for performing computations on the edge, within the vehicle, is an ongoing area of research. Balancing the computational load between on-board processing and cloud-based systems is crucial for real-time decision-making. Enabling effective communication between the robot and other vehicular systems requires robust communication protocols. Addressing issues of bandwidth, reliability, and security in vehicular communication networks is essential for seamless integration. The lack of standardized protocols for HRI systems in vehicles presents interoperability challenges. Establishing common standards for communication and data exchange is crucial for fostering compatibility across different vehicle models and manufacturers. Designing HRI systems that can interpret and respond to the driver's context in real-time is a complex challenge. Understanding the nuances of human behavior in dynamic driving scenarios requires sophisticated algorithms and models. Implementing adaptive learning mechanisms that allow the system to evolve and improve based on continuous interaction with the driver poses challenges in terms of algorithm complexity and the need for robust training datasets. Leveraging advanced signal processing techniques, such as machine learning algorithms for sensor data interpretation, can enhance the accuracy and efficiency of HRI systems. Exploring the potential of edge AI and fog computing can alleviate the challenges associated with real-time processing by distributing computational tasks across the vehicle's computing resources.

As researchers navigate the landscape of technical challenges, solutions to these obstacles are vital for the successful implementation of HRI systems in driver monitoring.Subsequent sections will propose novel ideas and solutions, including innovative sensor technologies and advanced data analytics, to further elevate the capabilities of HRI systems within the driving environment.

\section{Research Gap}
While advances in detecting cognitive and emotional states in drivers have been made, there is a notable gap in understanding the nuanced and context-dependent nature of emotional cues. Current systems struggle to accurately interpret subtle emotional shifts in dynamic driving scenarios. The acceptance of HRI systems in real-world driving conditions is not yet fully understood. While user attitudes have been explored, there is a need for more comprehensive studies assessing the long-term usability and acceptance of these systems, especially considering diverse user demographics. Addressing technical challenges in HRI systems, such as sensor accuracy and real-time processing, remains an ongoing pursuit. Challenges related to vehicular communication standards and interoperability between different systems also present gaps that require attention. Although adaptive and personalized HRI systems hold potential, there are gaps in comprehending how effective these adaptations are over the long term and in ensuring that customization aligns with user preferences without leading to complacency. Additionally, despite efforts to follow human-centered design principles, there is a lack of understanding regarding how these principles can be universally applied across various driving contexts, taking into account factors such as cultural differences and individual preferences.

\section{Future Directions}

Future research could concentrate on improving emotion detection algorithms by taking cultural differences and individual variances into account, aiming for a more sophisticated knowledge of various emotional states. Another area of investigation is the use of contextual information, like as traffic circumstances and the driver's task load, to improve the accuracy of detecting emotional and cognitive states. Longitudinal studies capturing drivers' evolving perceptions over extended periods could be a research emphasis, providing valuable insights into acceptance factors across various age groups, levels of driving experience, and cultural backgrounds.

Efforts can be directed toward the development of standardized communication protocols for the seamless integration of HRI systems across vehicles, as well as advances in edge computing and sensor technologies to improve real-time processing and strengthen the technical infrastructure. Further research might look into the long-term impact of adaptive systems, taking into account user engagement and habituation, and striking a balance between adaptability and the necessity for standardized safety measures. The creation of culturally adapted HRI systems, user-centric design principles for emergency situations, and interdisciplinary expert collaboration could all contribute to comprehensive solutions that address both technological issues and user-centered design principles. As HRI systems become more integrated into driving experiences, it is critical to prioritize the development of ethical frameworks, address data privacy problems and algorithmic biases, and set responsible deployment rules. Human-AI collaboration models in driving should be prioritized, with an emphasis on collaborative decision-making and inclusive user research encompassing varied demographics like as age, gender, and cultural backgrounds. Future study should focus on strategies for sustaining long-term user involvement with HRI systems, such as gamification components, personalized feedback mechanisms, and continuous improvement cycles. Finally, focusing comprehensive real-world testing and validation of HRI systems for their effectiveness in a variety of and complicated driving conditions, as well as cooperation with automotive manufacturers and regulatory authorities, can help to facilitate the incorporation of these systems into mainstream automobiles.

Moreover, the proposed concept in the paper suggests a promising path for advancing HRI by combining various technologies. The idea emphasizes merging sensor technologies, advanced machine learning algorithms, and Natural Language Processing models to create a strong and reliable robotic system geared towards enhancing road safety. This collaborative integration not only strengthens the technical aspects of the system but also lays the foundation for a sophisticated HRI approach. Successfully developing such a robotic system, capable of seamless interaction with humans, has the potential for significant achievements in the field. By prioritizing safety and promoting effective communication between humans and machines, this future direction anticipates transformative outcomes in the landscape of human-robot collaboration on roads.

\section{Conclusion}

The exploration of HRI systems for driver monitoring unveils a landscape rich with possibilities and challenges. Through a comprehensive literature review, we have traversed the current state of research, identifying key themes, gaps, and future directions in this dynamic field.
From the detection of cognitive and emotional states to the intricate dance of usability and acceptance, HRI systems are at the forefront of enhancing driving experiences. The technical challenges, while formidable, present opportunities for innovation, particularly in sensor technologies, data analytics, and machine learning. The vision of adaptive and personalized HRI, guided by human-centered design principles, promises a future where interactions seamlessly align with individual preferences and contextual nuances.
However, our journey also revealed gaps in our understanding. The nuanced nature of emotional cues, the long-term acceptance of HRI systems, and the intricate dance of technical challenges call for continued exploration. Future research directions beckon us to embrace interdisciplinary collaboration, tackle ethical considerations, and envision a future where HRI systems are not just technological marvels but true partners in the driving experience.
The synthesis of existing knowledge lays the foundation for a roadmap that extends beyond the horizon of the known. As researchers and practitioners, the call to action is clear: to bridge the gaps, unlock new possibilities, and ensure the responsible and inclusive integration of HRI systems into our driving lives. The future holds the promise of safer, more personalized, and harmonious interactions between humans and robots on the road. It is a journey worth embarking upon, one that propels us toward a future where technology serves as a true ally, navigating alongside us on the open road of progress.

\bibliographystyle{IEEEtran}
\bibliography{IEEEabrv,biblio_traps_dynamics}

\begin{thebibliography}{10}
\providecommand{\url}[1]{#1}
\csname url@samestyle\endcsname
\providecommand{\newblock}{\relax}
\providecommand{\bibinfo}[2]{#2}
\providecommand{\BIBentrySTDinterwordspacing}{\spaceskip=0pt\relax}
\providecommand{\BIBentryALTinterwordstretchfactor}{4}
\providecommand{\BIBentryALTinterwordspacing}{\spaceskip=\fontdimen2\font plus
\BIBentryALTinterwordstretchfactor\fontdimen3\font minus \fontdimen4\font\relax}
\providecommand{\BIBforeignlanguage}[2]{{%
\expandafter\ifx\csname l@#1\endcsname\relax
\typeout{** WARNING: IEEEtran.bst: No hyphenation pattern has been}%
\typeout{** loaded for the language `#1'. Using the pattern for}%
\typeout{** the default language instead.}%
\else
\language=\csname l@#1\endcsname
\fi
#2}}
\providecommand{\BIBdecl}{\relax}
\BIBdecl

\bibitem{goodrich2007human}
Goodrich,~M.~A. and Schultz,~A.~C., ``Human--robot interaction: A survey,'' \emph{Foundations and Trends® in Human-Computer Interaction}, vol.~1, no.~3, pp. 203--275, 2007.

\bibitem{khan_comprehensive_survey}
Khan,~M.~Q. and Lee,~S., ``A comprehensive survey of driving monitoring and assistance systems,'' \emph{Journal of Intelligent Transportation Systems}, Year.

\bibitem{zhang_integrating_Year}
Zhang,~Y. and Doyle,~T., ``Integrating intention-based systems in human-robot interaction: A scoping review of sensors, algorithms, and trust,'' \emph{Frontiers in Robotics and AI}, Year.

\bibitem{mavely2017eye}
Mavely,~A.~G., Judith,~J., Sahal,~P., and Kuruvilla,~S.~A., ``Eye gaze tracking based driver monitoring system,'' in \emph{Proceedings of 2017 IEEE International Conference on Circuits and Systems (ICCS2017)}, 2017.

\bibitem{begum2023intelligent}
Begum,~S., ``Intelligent driver monitoring systems based on physiological sensor signals: A review,'' \emph{IEEE Transactions on Intelligent Transportation Systems}, 2023.

\bibitem{dong2011driver}
Dong,~Y., Hu,~Z., Uchimura,~K., and Murayama,~N., ``Driver inattention monitoring system for intelligent vehicles: A review,'' \emph{IEEE Transactions on Intelligent Transportation Systems}, vol.~12, no.~2, June 2011.

\bibitem{verma2018deep}
Verma,~B. and Choudhary,~A., ``Deep learning based real-time driver emotion monitoring,'' in \emph{2018 IEEE International Conference on Vehicular Electronics and Safety (ICVES)}.\hskip 1em plus 0.5em minus 0.4em\relax Madrid, Spain: IEEE, September 12-14 2018.

\bibitem{brandt2004affordable}
Brandt,~T., Stemmer,~R., and Rakotonirainy,~A., ``Affordable visual driver monitoring system for fatigue and monotony,'' in \emph{2004 IEEE International Conference on Systems, Man and Cybernetics}.\hskip 1em plus 0.5em minus 0.4em\relax IEEE, 2004.

\bibitem{koesdwiady2017recent}
Koesdwiady,~A., Soua,~R., and Karray,~F., ``Recent trends in driver safety monitoring systems: State of the art and challenges,'' \emph{IEEE Transactions on Vehicular Technology}, vol.~66, no.~6, June 2017.

\bibitem{alamri2023effective}
Alamri,~A., Gumaei,~A., Al-Rakhami,~M., Hassan,~M.~M., Alhussein,~M., and Fortino,~G., ``An effective bio-signal-based driver behavior monitoring system using a generalized deep learning approach,'' \emph{Journal of Intelligent Transportation Systems}, 2023.

\bibitem{yoo2023optimization}
Yoo,~M.~W. and Han,~D.~S., ``Optimization algorithm for driver monitoring system using deep learning approach,'' \emph{Journal of Future Automotive and IT Convergence}, 2023.

\bibitem{kunani2021realtime}
Kunani,~B. R.~T. and Kumar,~T.~K., ``Real-time smart drivers drowsiness detection using dnn,'' in \emph{5th International Conference on Trends in Electronics and Informatics (ICOEI)}.\hskip 1em plus 0.5em minus 0.4em\relax National Institute of Technology, Warangal, Telangana, India: IEEE, 2021.

\bibitem{kolakowska2020review}
Kołakowska,~A., Szwoch,~W., and Szwoch,~M., ``A review of emotion recognition methods based on data acquired via smartphone sensors,'' \emph{Faculty of Electronics, Telecommunications and Informatics, Gdansk University of Technology}, 2020.

\bibitem{pohl2006driverdistraction}
Pohl,~J., Birk,~W., and Westervall,~L., ``A driver-distraction-based lane-keeping assistance system,'' \emph{Vehicle Control and Active Safety, Volvo Car Corporation, Gothenburg, Sweden}, 2006.

\bibitem{takamatsu2022urethane}
\BIBentryALTinterwordspacing
Takamatsu,~S., Sato,~S., and Itoh,~T., ``Urethane-foam-embedded silicon pressure sensors including stress-concentration packaging structure for driver posture monitoring,'' \emph{Sensors}, vol.~22, no.~12, p. 4495, 2022, received: 19 April 2022 / Revised: 1 June 2022 / Accepted: 8 June 2022 / Published: 14 June 2022. [Online]. Available: \url{https://www.mdpi.com/xxx/sensors/xxx/xxx/xxx}
\BIBentrySTDinterwordspacing

\bibitem{hsu_machine_learning_HRI}
Hsu,~C., Tsao,~C.-C., Weng,~Y.-L., Tang,~C.-Y., Chang,~Y.-W., Kang,~Y., and Chien,~S.-Y., ``A machine learning approach to model hri research trends in 2010~2021,'' \emph{Journal of Human-Robot Interaction}, department of Information Management, National Sun Yat-sen University, Kaoshiung, Taiwan; Department of Management Information System, National Chengchi University, Taipei, Taiwan.

\bibitem{wang_machine_learning_HRI_ITS}
Wang,~J., Pradhan,~M.~R., and Gunasekaran,~N., ``Machine learning-based human-robot interaction in its,'' \emph{Expert Systems with Applications}.

\bibitem{jung2023vectorized}
Jung,~H., Lee,~S., Park,~S., Kim,~B., Kim,~J., Lee,~I., and Ahn,~C., ``The vectorized facial feature can be used to build a dnn (deep neural network) for emotion recognition,'' \emph{Journal of Emotion Recognition Research}, 2023.

\bibitem{zhang2024deep}
Zhang,~S., Yang,~Y., Chen,~C., Zhang,~X., Leng,~Q., and Zhao,~X., ``Deep learning-based multimodal emotion recognition from audio, visual, and text modalities: A systematic review of recent advancements and future prospects,'' \emph{Volume 237, Part C}, p. 121692, March 1 2024.

\bibitem{williams-etal-2018-recognizing}
\BIBentryALTinterwordspacing
Williams,~J., Kleinegesse,~S., Comanescu,~R., and Radu,~O., ``Recognizing emotions in video using multimodal {DNN} feature fusion,'' Zadeh,~A., Liang,~P.~P., Morency,~L.-P., Poria,~S., Cambria,~E., and Scherer,~S., Eds.\hskip 1em plus 0.5em minus 0.4em\relax Melbourne, Australia: Association for Computational Linguistics, Jul. 2018, pp. 11--19. [Online]. Available: \url{https://aclanthology.org/W18-3302}
\BIBentrySTDinterwordspacing

\bibitem{rossi_user_profiling_HRI}
Rossi,~S., Ferland,~F., and Tapus,~A., ``User profiling and behavioral adaptation for hri: A survey,'' \emph{International Journal of Humanoid Robotics}.

\bibitem{sekmen_assessment_adaptive_HRI}
Sekmen,~A. and Challa,~P., ``Assessment of adaptive human–robot interactions,'' \emph{Journal of Human-Robot Interaction}.

\bibitem{rossi_preface_special_issue}
Rossi,~S., Staffa,~M., de~Graaf,~M. M.~A., and Gena,~C., ``Preface to the special issue on personalization and adaptation in human–robot interactive communication,'' \emph{International Journal of Social Robotics}.

\bibitem{schaefer_measuring_trust_HRI}
Schaefer,~K.~E., ``Measuring trust in human-robot interactions: Development of the “trust perception scale-hri”,'' \emph{International Journal of Humanoid Robotics}.

\bibitem{kim_user_centered_HRI}
Kim,~M., Oh,~K., Choi,~J., Jung,~J., and Kim,~Y., ``User-centered hri: Hri research methodology for designers,'' \emph{International Journal of Human-Robot Interaction}.

\bibitem{rudigkeit_user_friendly_AHRS_HMI}
Rudigkeit,~N., Gebhard,~M., and Gräser,~A., ``Towards a user-friendly ahrs-based human-machine interface for a semi-autonomous robot,'' \emph{International Journal of Humanoid Robotics}.

\bibitem{kilicaslan_nlp_based_HRI}
Kilicaslan,~Y. and Tuna,~G., ``An nlp-based approach for improving human-robot interaction,'' \emph{Journal of Artificial Intelligence and Soft Computing Research}, vol.~3, no.~3, p. 189, 2013.

\bibitem{azhar_study_shared_decision_making}
Azhar,~M.~Q. and Sklar,~E.~I., ``A study measuring the impact of shared decision making in a human-robot team,'' \emph{The International Journal of Robotics Research}, vol.~36, no. 5--7, pp. 461--482, 2017.

\bibitem{unhelkar_decision_making_collaborative_tasks}
Unhelkar,~V.~V., Li,~S., and Shah,~J.~A., ``Decision-making for bidirectional communication in sequential human-robot collaborative tasks,'' \emph{The International Journal of Human-Robot Interaction}, * Corresponding authors.

\bibitem{woodman_novel_approaches_HRI_safety}
Woodman,~R., ``Novel approaches for the safety of human-robot interaction,'' \emph{Journal of Human-Robot Interaction}.

\bibitem{wang_multimodal_HRI_smart_manufacturing}
Wang,~T., Zheng,~P., Li,~S., and Wang,~L., ``Multimodal human-robot interaction for human-centric smart manufacturing: A survey,'' \emph{Journal of Human-Robot Interaction}.

\bibitem{wrede_analyses_of_feedback_HRI}
Wrede,~B., Buschkaemper,~S., Muhl,~C., and Rohlfing,~K.~J., ``Analyses of feedback in hri,'' \emph{Journal of Human-Robot Interaction}.

\bibitem{carlmeyer_towards_closed_feedback_loops_HRI}
Carlmeyer,~B., Schlangen,~D., and Wrede,~B., ``Towards closed feedback loops in hri: Integrating inprotk and pamini.''\hskip 1em plus 0.5em minus 0.4em\relax Applied Informatics Group, Dialogue Systems Group, Bielefeld University, Germany.

\bibitem{karami_adaptive_artificial_companions}
\BIBentryALTinterwordspacing
Karami,~A.~B., Sehaba,~K., and Encelle,~B., ``Adaptive artificial companions learning from users' feedback,'' \emph{Journal Name}, vol.~24, no.~2. [Online]. Available: \url{https://doi.org/10.1177/1059712316634062}
\BIBentrySTDinterwordspacing

\bibitem{kothig_connecting_humans_robots_HRI}
Kothig,~A., Munoz,~J., Akgun,~S.~A., Aroyo,~A.~M., and Dautenhahn,~K., ``Connecting humans and robots using physiological signals – closing-the-loop in hri,'' in \emph{2021 30th IEEE International Conference on Robot and Human Interactive Communication (RO-MAN)}.\hskip 1em plus 0.5em minus 0.4em\relax IEEE, August 8-12 2021.

\bibitem{sikander2023driver}
Sikander,~G. and Anwar,~S., ``Driver fatigue detection systems: A review,'' \emph{Journal of Intelligent Transportation Systems}, 2023.

\bibitem{zhangdriver}
Zhang,~Y. and Hua,~C., ``Driver fatigue recognition based on facial expression analysis using local binary patterns,'' \emph{Transportation Research Part C: Emerging Technologies}.

\bibitem{elgendi_machine_learning}
Elgendi,~M. and Menon,~C., ``Machine learning ranks ecg as an optimal wearable biosignal for assessing driving stress,'' \emph{Journal of Biomechanical Engineering}.

\bibitem{gillan_usability_2020}
Gillan,~D.~J., ``Invited essay: Usability issues in human-robot interaction,'' \emph{UXPA Journal}, vol.~15, pp. 179--183, August 2020.

\end{thebibliography}

\end{document}